\pdfoutput=1

\documentclass[11pt]{article}

\usepackage[]{naacl2021}

\usepackage{times}
\usepackage{latexsym}

\usepackage[T1]{fontenc}

\usepackage[utf8]{inputenc}

\usepackage{microtype}
\usepackage{graphicx}
\usepackage{multirow}
\usepackage{amsmath}

\title{Multi-Style Transfer with Discriminative Feedback on Disjoint Corpus}

\author{Navita Goyal, Balaji Vasan Srinivasan, Anandhavelu N, Abhilasha Sancheti \\
  Adobe Research, India \\
  \{navgoyal, balsrini, anandvn, sancheti\}@adobe.com}

\begin{document}
\maketitle
\begin{abstract}
Style transfer has been widely explored in natural language generation with non-parallel corpus by directly or indirectly extracting a notion of style from source and target domain corpus. A common shortcoming of existing approaches is the prerequisite of joint annotations across all the stylistic dimensions under consideration. Availability of such dataset across a combination of styles limits the extension of these setups to multiple style dimensions. While cascading single-dimensional models across multiple styles is a possibility, it suffers from content loss, especially when the style dimensions are not completely independent of each other. In our work, we relax this requirement of jointly annotated data across multiple styles by using independently acquired data across different style dimensions without any additional annotations. We initialize an encoder-decoder setup with transformer-based language model pre-trained on a generic corpus and enhance its re-writing capability to multiple target style dimensions by employing multiple style-aware language models as discriminators. Through quantitative and qualitative evaluation, we show the ability of our model to control styles across multiple style dimensions while preserving content of the input text. We compare it against baselines involving cascaded state-of-the-art uni-dimensional style transfer models.
\end{abstract}

\section{Introduction}
Style transfer is a popular task in natural language processing and has been studied on attributes like age or gender \cite{subramanian-2018-multiple-attribute}, styles emanating from social construct like formality \cite{rao-tetreault-2018-gyafc} and politeness \cite{madaan2020politeness}, linguistic styles based on author writing style \cite{adapting-2020-rewritting}, or psycho-linguistic styles based on personality types \cite{mairesse-walker-2011-controlling}. While early style transfer frameworks were modeled as a supervised learning task on a parallel corpus, state-of-the-art models are semi-supervised/unsupervised and operate on non-parallel corpus. These models achieve style transfer by aligning source and target distribution of sentences from non-parallel corpus \cite{shen-2017-cross}, disentangling content space from style space in latent representation \cite{hu-2017-controlled} or employing self-reconstruction \cite{dai-etal-2019-style} and back translation \cite{lample-etal-2018-dae} objectives to achieve pseudo-supervision with non-parallel corpus. Recent works have also modeled this in a self-supervised manner where rewriting (transfer) is achieved by utilizing corpus from the target style alone \cite{adapting-2020-rewritting}. These wide studies have also led to the curation and benchmarking of non-parallel dataset for various style dimensions, such as sentiment \cite{li-etal-2018-delete}, formality \cite{rao-tetreault-2018-gyafc}, politeness \cite{danescu-niculescu-mizil-etal-2013-stanford-politeness}, excitement \cite{sancheti-2020-excietement}, etc. But availability of data with joint tagging across multiple styles is limited and has restricted the ability of existing approaches to scale from single-dimensional transfer to multiple style dimensions. In this paper, we propose a multi-dimensional style transfer approach that can work off partially labelled data for style transfer across multiple dimensions simultaneously. 

The work by \newcite{subramanian-2018-multiple-attribute} attempts style transfer with multiple attributes such as age, gender, and sentiment simultaneously. However, their approach avails corpus tagged with each of these three style dimensions. 
In contrast to this and other similar explorations in multi-style transfer, our approach does not require jointly labelled data across all the stylistic dimensions in source and/or target corpus. We focus on the problem where independent corpus is available across different stylistic dimensions (say \textit{sentiment} and \textit{formality}) and we achieve style transfer spanning different stylistic dimensions (say make a sentence more \textit{positive} and \textit{formal}). While state-of-the-art approaches can be extended to achieve this by sequentially transferring one style after another, it is limited as different style dimensions are not necessarily independent of each other. In aspects that are not independent, changing one style aspect of the text might affect another aspect considered, making a sequential brute-force approach non-ideal. As we show in our experiments later, the cascaded setup also lacks common grounding between the content from different styles leading to erratic changes in content. We circumvent this by grounding our framework on the linguistic understanding of a large language model. Our model builds understanding of interplay between the different styles by incorporating multiple discriminative language models (LM) with language model-based encoder-decoder setup. The key contributions of this paper are: 

\noindent $1)$ An encoder-decoder setup with multiple language models as discriminator, with each entity harnessing the language understanding from a large pre-trained transformer model.

\noindent $2)$ Relaxing the requirement of jointly labelled data for multi-style transfer, by leveraging independently acquired disjoint corpus for different styles.

\noindent $3)$ Achieving better style control with better content preservation in multi-dimensional style transfer than a cascaded setup of state-of-the-art uni-dimensional style transfer models.

\section{Related Work}
One line of work in \textbf{style transfer} attempts to learn disentangled latent representation for style and content, and transfer style by manipulating latent representation of style \cite{shen-2017-cross}. Although these approaches perform well with one style at a time, they do not trivially scale to multi-dimensional style transfer. Several other works develop unsupervised approach for style transfer by employing Denoising Autoencoding (DAE) \cite{fu-etal-2017-exploration} and back-translation (BT) \cite{lample-etal-2018-dae} loss to develop interaction and hence transfer between the source and target domain. \newcite{subramanian-2018-multiple-attribute} extend this approach to multiple styles by conditioning on average of embedding of each target attribute and using combination of DAE and back-translation techniques. DAE takes as input a sentence $x$ from style $s$ and tries to reconstruct sentence $x$ from its corrupted version $\tilde{x}$. This relies on the assumption that the input sentence $x$ is from a certain style combination $s=\{s_1, s_2, \ldots, s_k\}$. Similarly back translation (BT) objective with input sentence $x$ from style $s$, first estimates $x'=f(x, s')$, where $s\neq s'$ and then reconstruct $x$ from $\tilde{x}=f(x',s)$.
Thus, these approaches are inherently dependent on knowledge of annotation of each sentence with all the style combinations. 
\newcite{dai-etal-2019-style} achieve state-of-the-art style transfer in single style dimensions by employing transformer-based model in conjunction with classifier-based discriminator. In addition to discriminator losses, their proposed technique uses self-reconstruction and cycle reconstruction losses, which similar to DAE and BT losses are also reliant on availability of jointly annotated data to be extendable to multiple style setup. 

\textbf{Language modeling} is integral to several natural language generation (NLG) tasks like text summarization, spelling correction, image captioning, etc. The model architecture for these tasks has evolved from n-gram based methods to Recurrent Neural Networks to transformer architectures. The introduction of Transformer-based architecture accompanied with generative pre-training \cite{radford-2018-gpt} capabilities have led to strong improvements in many downstream generation and GLUE \cite{wang-etal-2018-glue} tasks. Generative pre-training aims to adapt a large Transformer language model to large unsupervised corpus. This capability of generative pre-training is exploited in many large language models like BERT \cite{devlin-etal-2019-bert}, GPT-2 \cite{radford-wu-2018-gpt-2}, ERNIE 2.0 \cite{sun-wang-2019-ernie} which have the ability to perform tasks like reading comprehension \cite{xu-etal-2019-bert-reading}, summarization \cite{liu-lapata-2019-bert-summary}, question-answering \cite{rajpurkar-etal-2016-squad} and translation \cite{clinchant-etal-2019-bert-nmt} in zero-shot and few-shot settings.

Recently these pre-trained generative language models have been explored in translation \cite{lample-2019-cross-lingual} and style transfer tasks \cite{adapting-2020-rewritting}. \newcite{lample-2019-cross-lingual} develop cross-lingual models for unsupervised machine translation by initializing encoder and decoder with a pre-trained language model trained on Masked Language Modeling (MLM) \cite{devlin-etal-2019-bert} objective and fine-tuning the encoder-decoder framework with adversarial training. \newcite{adapting-2020-rewritting} extend this to stylized re-writing task by employing DAE during fine-tuning. The joint encoder-decoder framework learns to reconstruct sentences in target-domain from its noisy version using DAE objective. As previously discussed, the DAE objective is reliant on the corpus being tagged for the target domain style (or combination of style) and restricts the generalization of this setup to multiple attributes.
We overcome this by employing discriminative language models to assist the decoder with feedback for various target styles. 

\newcite{shen-2017-cross} show that even with non-parallel data, the content distribution across source and target style is shared. Based on this, a language model trained on target style will have high perplexity on transferred text if it does not match target style and low perplexity otherwise. \newcite{NIPS2018_7959} exploit this ability of language models to replace standard binary classifier-based discriminators with an implicitly trained language model as discriminator. They show that using the language model as structured discriminator allows for more stable training by eliminating the adversarial step.
We extend this idea to a multi-discriminator approach. Training a LM on combination of target styles is not possible in absence of jointly labelled dataset. Due to this, we attempt to use multiple discriminators for each of the target styles. Since with multiple styles, the underlying corpus is independently acquired, the variation in content distribution across different styles is more noticeable. Consequently, an independently trained LM on one of the target styles might have high perplexity even if the transferred sentence fits in the corresponding target style, due to the content space of source sentence. To equip discriminative LM with more generalized notion of content, we use large transformer-based LM pre-trained on large unsupervised corpus to establish generic content distribution before style-oriented fine-tuning.

\begin{figure*}[t]
\centering
\includegraphics[width=0.9\textwidth]{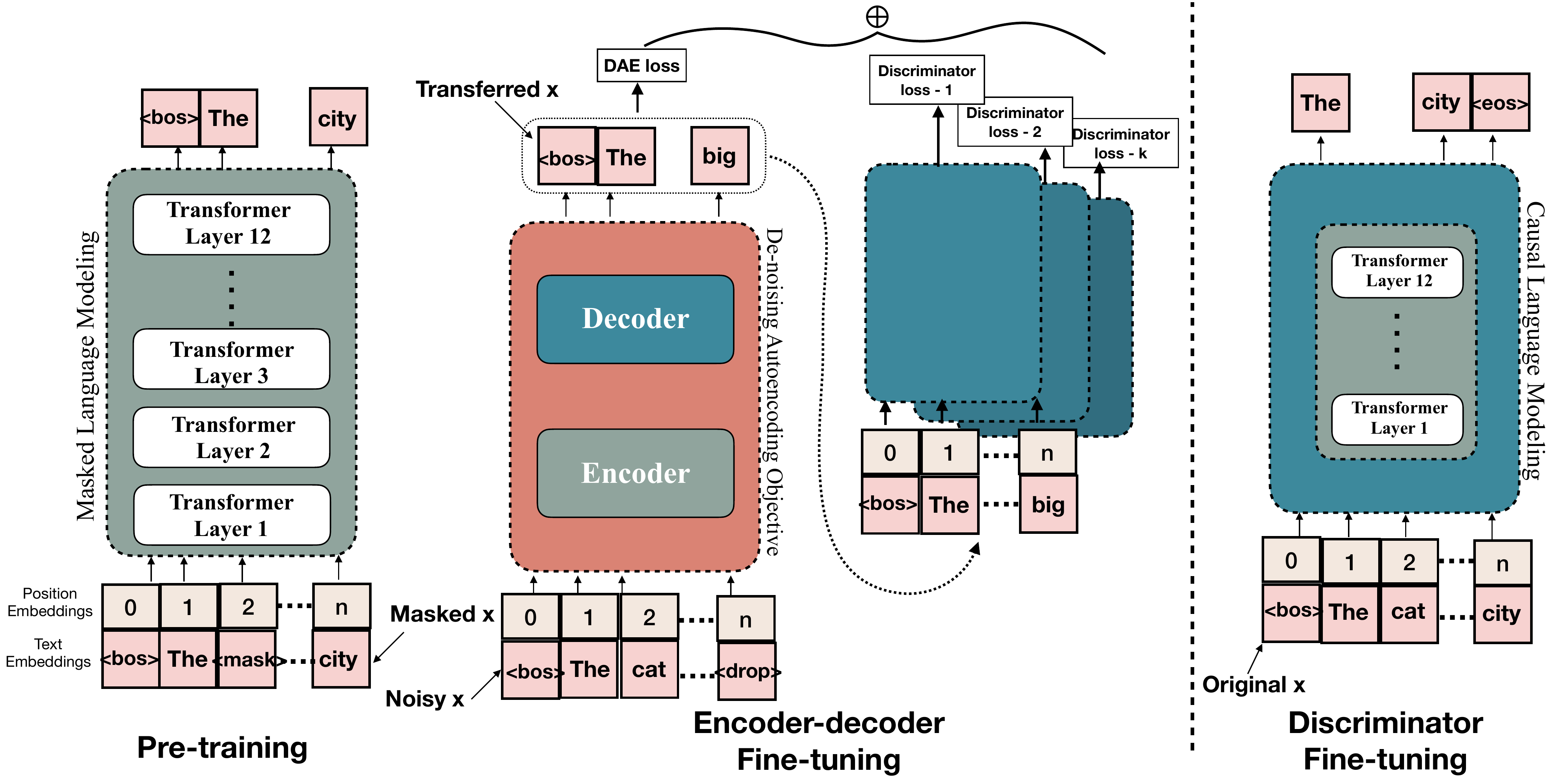} 
\caption{Model Architecture - Left: Generative pre-training using MLM objective, and Fine-tuning encoder-decoder LM with multiple discriminative losses and Right: Discriminator fine-tuning with language modeling (next token prediction) objective. Color for model blocks represents the pre-trained model used for initialization prior to fine-tuning.
}
\label{fig:model_architecture}
\end{figure*}

\section{Approach}
Our proposed approach has two key elements --- a Transformer-based encoder-decoder model initialized with a pre-trained Transformer Language Model and fine-tuned on DAE loss to achieve style transfer (Section~\ref{section-cascaded-lm}) and the multiple language models as discriminators stacked together to enable multi-style transfer (Section~\ref{section-LM-disc}).

\subsection{Pre-trained LM as Encoder-Decoder}
\label{section-cascaded-lm}
Similar to \newcite{adapting-2020-rewritting}, we first pre-train a Transformer-based language model with Masked Language Modeling (MLM) objective on English Wikipedia data extracted using WikiExtractor.\footnote{https://github.com/attardi/wikiextractor} This equips LM with the ability to predict masked words over a large corpus. Masked Language Modeling leverages bidirectional context of the input, thus enabling better language understanding. Following Masked Language Modeling objective from \newcite{devlin-etal-2019-bert}, we randomly sample $15\%$ of the tokens from the text stream and replace them with the [MASK] token $80\%$ of the time, by a random token $10\%$ of the time and keep them unchanged $10\%$ of the time, with the objective of predicting the original identity of the masked word based on its bidirectional context. 
To enable style transfer from a given sentence to target style, we use independently trained language models (LMs) to initialize the encoder and decoder and connect these with randomly initialized attention layers to arrive at a \textit{encoder-decoder setup}. 
As discussed by \newcite{adapting-2020-rewritting}, the Transformer architecture \cite{vaswani-2017-attention} allows such independent initialization by implicitly aligning encoder-decoder layers via attention mechanism. 

Pre-training an encoder only transformer on generative task and then leveraging it to initialize as both encoder and decoder as opposed to pre-training a joint encoder-decoder model has several advantages. Transformer-based models with encoder-only \cite{devlin-etal-2019-bert} or decoder-only \cite{radford-wu-2018-gpt-2} blocks have been shown to perform well in generative pre-training task. Clearly, pre-training a single transformer block on generative task and then utilizing it as both encoder and decoder blocks has lower computational cost than training the entire encoder-decoder block jointly. 
Moreover, this also enables us to use the same pre-trained model to initialize both style transfer module and the discriminator models, explained in the following section. This is not only computationally more efficient but it also closely ties the underlying language distribution of the two modules. This is expected to make the discriminative feedback more effective while fine tuning the transfer model for multiple styles.

In \newcite{adapting-2020-rewritting}'s setup, both encoder and decoder in the style transfer module are initialized with the pre-trained language model (trained on MLM objective). Instead, we initialize the decoder with the language model fine-tuned with the target style using Causal Language Modeling (CLM) objective, before training the joint encoder-decoder model, as detailed in Section~\ref{section-LM-disc}. The encoder is initialized with the pre-trained model directly. Aligning the decoder to the distribution of the target style helps speed up the fine-tuning process as decoder is more adept at generating stylized outputs. This does not add to computational overhead as these fine-tuned models are repurposed as discriminators for stylistic feedback (Section~\ref{section-LM-disc}). 

To instill style-awareness to the encoder-decoder setup initialized with pre-trained Transformer models, we fine-tune it with Denoising Autoencoder (DAE) loss using the target-domain corpus. In case of multiple styles, we use a randomized mixture of target-domain corpus from each of the target styles.
Under the DAE objective, the encoder takes a noisy masked version $\tilde{x}$ of the text $x$ as input and attempts to fill in the mask token as per the MLM objective that it was pre-trained on. In turn, the decoder re-creates stylistic version of original sentence from this noisy output from the encoder. The overall training objective is
\begin{equation}
\label{eq:gen_loss}
    \mathcal{L}_{DAE}(\theta_G) = \mathbf{E}_{x \sim T}[-\log P_{\theta_G}(x|\tilde{x})],
\end{equation}
where $\theta_G$ are the trainable parameters of the encoder-decoder model. The noisy version of sentence $x$ from the target corpus $T$ is obtained after dropping tokens from $x$ with probability $p_{drop}$ and masking with a probability of $p_{mask}$.
In conjunction, the encoder and decoder enable style transfer to the target style. The noteworthy aspect here is that the model has no sense of source style and is trained to generate sentences to match the style of the target-domain corpus with which it is trained.

\subsection{Fine-tuned LM as discriminators}
\label{section-LM-disc}
To extend the single-dimensional style transfer setup above to multi-dimensional setting, we use language models as discriminators to provide the feedback to the model for partially annotated nature of input data. As opposed to a classifier-based discriminator, the language model as discriminator takes into account the wider language distribution of the target style. Additionally, such a setup allows us to use only the target style corpus for training the transfer model, whereas the classifier would require both source and target style corpus to distinguish between a sentence as being from one style or another.
Inspired by \newcite{NIPS2018_7959}, we fine-tune a language model on the target style $s_i$, so that the language model is equipped with language distribution of target domain data. 
This entails generating the probability of next token, given the previous tokens --- also known as Causal Language Modeling objective \cite{lample-2019-cross-lingual}. The training loss for the LM for target style $s_i$ with corresponding corpus $T_i$ is 
\begin{equation}\label{eq:clm_loss}
\small
    \mathbf{E}_{x \sim T_i}\bigg[\sum_{t=1}^n[-\log P_{LM}(x_t|x_1,\ldots, x_{t-1})]\bigg]
\end{equation}

We show in our experiments that such a fine-tuning step transforms language distribution of this language model to style $s_i$ and hence serve as soft-discriminator for our framework. We exploit this capability of language models to imbibe style of fine-tuning corpus by employing language models as style discriminators for transferred sentences. 
This is based on the idea that if the transferred sentence does not fit well in the target style, then the perplexity of language model fine-tuned on that style will be high (Section~\ref{section-lm-experiments}).

For $k$-dimensional style transfer with target styles $s=\{s_1, s_2, \ldots, s_k\}$, we independently fine-tune $k$ language models on each of the target styles. As discussed in \newcite{NIPS2018_7959}, we are able to forgo the adversarial training for the discriminator, since the fine-tuned discriminative language model is implicitly capable of assigning high perplexity to negative samples (out-of-style samples), as shown in Section~\ref{section-lm-experiments}.
For the transferred sentence $x'$, the training objective for each target style $s_i$ is,
\begin{equation}
\begin{split}
    \underset{\theta_G}{\operatorname{arg min}}\mathcal{L}^{s_i} &= \mathbf{E}_{x\sim T, x'\sim P_{\theta_G}(x)}\\
    \bigg[ &\sum_{t=1}^n -\log P_{{LM}_i}(x'_t|x'_1,.., x'_{t-1})\bigg]
\end{split}
\end{equation}
This dictates that transferred sentence $x'$ has low perplexity on the language model fine-tuned on style $s_i$, for each target style $s_i$. However, we cannot directly find the $\operatorname{argmin}_{\theta_G}$ using gradient descent because of discrete sampling of $x'\sim P_{\theta_G}(x)$. To account for this, we use a policy gradient reinforcement learning approach using REINFORCE algorithm \cite{sutton-1999-reinforce}. The reward for an input sequence $x$ to the style discriminator $LM_i$ is calculated as,
\begin{equation}
\label{eq:reward}
    r(x)=\sum_{t=1}^n \log P_{{LM}_i}(x_t|x_1,.., x_{t-1})
\end{equation}
Using these rewards, the RL objective is to minimize the loss $\mathcal{L}^{s_i}$ given by,
\begin{equation}
\label{eq:disc_loss}
    \begin{split}
        \mathcal{L}^{s_i} = \mathbf{E}_{x\sim T, x'\sim P_{\theta_G}(x)}
        &(r(x')-r(x))\\
        &[-\log P_{\theta_G}(x'|\tilde{x})]
    \end{split}
\end{equation}
for style $s_i$, where $P_{\theta_G}(x|\tilde{x})$ is as in Equation~\ref{eq:gen_loss} and $r(x')$ is the reward in the Equation~\ref{eq:reward} for the transferred sentence $x'$. The rewards $r(x)$ represents the baseline reward of greedily sampling the input sequence $x$ by the style discriminator $LM_i$.

For the style combination $s=\{s_1, s_2, \ldots, s_k\}$, the joint encoder-decoder model is trained on randomized mixture of data from each of the target-domain corpus.  The mixture is thus agnostic of individual style of each of the sentence and the discriminative LM for each style guides the generation towards that specific style by rewarding style adherence in the transferred sentence. Randomized mixture of training corpus across styles allows for unified and cohesive understanding of multiple styles by diversifying rewards from different discriminators across samples. The overall training loss for the joint encoder-decoder model is
\begin{equation}\label{eq:dae_w_disc_loss}
\small
    \mathcal{L} =\lambda_{DAE} \mathbf{E}_{x \sim T}[-\log P_{\theta}(x|\tilde{x})]+\sum_{i=1}^k \lambda_i \mathcal{L}^{s_i},
\end{equation}
where $\mathcal{L}^{s_i}$ is as defined in Equation~\ref{eq:disc_loss}, and $\lambda_{DAE}$ and $\{\lambda_i\}_{i=1}^k$ are hyper-parameters. 

The overall training process is summarized in \figurename~\ref{fig:model_architecture}. First, we pre-train a transformer model with Masked language modeling objective as shown in \figurename~\ref{fig:model_architecture}(Left). We then initialize discriminator model with this pre-trained language model and fine-tune it with Causal language modeling objective, shown in \figurename~\ref{fig:model_architecture}(Right), for each target style. Finally, we initialize the encoder and decoder of the style transfer module with the pre-trained and style-specific fine-tuned language models, respectively. In case of multiple styles, the decoder can be initialized with the language model which is fine-tuned with CLM loss on the mixture of data from target styles, i.e., CLM loss in Equation~\ref{eq:clm_loss} with $x\sim T$. The joint encoder-decoder model (\figurename~\ref{fig:model_architecture}(Centre)) is then trained with a combination of DAE objective and rewards from fine-tuned discriminators of respective target styles. 

\section{Experiments}
We experiment with a combination of sentiment and formality styles. For sentiment, we use a mixture of IMDB \cite{maas-EtAl:2011:ACL-HLT2011} and Yelp dataset \cite{li-etal-2018-delete} with $300k$ examples in the positive and negative sentiment each. For formality, we use GYAFC corpus \cite{rao-tetreault-2018-gyafc} which has $104k$ examples in each formal and informal class. The test set has $3000$ and $4849$ examples for sentiment and formality respectively, following the data split available in \citet{dai-etal-2019-style, rao-tetreault-2018-gyafc}.
For both datasets, the training corpus is non-parallel and the test corpus has human written references available, which we use for content evaluation (Section~\ref{section-evaluation}). 

For pre-training, we use $12$-layer Transformer model with $512$ hidden units, $16$ heads, a dropout rate of $0.1$ and
learned positional embedding. We train our models with the
Adam optimizer, and a learning rate of $10^{-4}$. To handle large vocabulary sizes, we use Byte Pair Encoding (BPE) \cite{sennrich-etal-2016-bpe} learned on the Wikipedia dataset. The $\lambda$s in Equation~\ref{eq:dae_w_disc_loss} are determined using hyper-parameter tuning on validation set, with style transfer accuracy (Section~\ref{section-evaluation}) as search criteria.

\begin{table}[t]
    \centering
    \begin{tabular}{l|cc}
        \hline
        \textbf{Style/Dimension} & \textbf{Sentiment} $\%$ & \textbf{Formality} $\%$ \\
        \hline
        Positive & 71.41 & 67.09 \\
        Negative & 76.17 & 75.59 \\
        \hline
    \end{tabular}
    \caption{Accuracy of sentences generated by model fine-tuned on style $s_i$ as \% of generated sentences labelled as class $s_i$ by the classifier trained on the corresponding style dimension. }
    \label{table-style-fine-tune}
\end{table}

\begin{table}[t]
    \centering
    \begin{tabular}{l|cc}
        \hline
        \textbf{Fine-tuning} & \multicolumn{2}{|c}{\textbf{Test Corpus}} \\
        \cline{2-3}
        \textbf{corpus} & Same $\downarrow$ &  Opposite $\uparrow$\\
        \hline
        Positive & 6.9275 & 9.6850 \\
        Negative & 7.7131 & 9.9637 \\
        \hline
    \end{tabular}
    \caption{Perplexity of test corpus on models fine-tuned positive and negative corpus (rows). The column \textit{Same} represents that test corpus is same as fine-tuning corpus, leading to lower perplexities and  \textit{Opposite} represent test corpus from opposite polarity as fine-tuning corpus leading to higher perplexity.}
    \label{table-fine-tuning-perplexity}
\end{table}

\begin{table*}[t]
\centering
\small
\begin{tabular}{c|cc|c|cc|c}
\hline
\multirow{3}{*}{\textbf{Model}} & \multicolumn{3}{|c|}{\textbf{Style Accuracy}} & \multicolumn{2}{|c|}{\textbf{Content Preservation}} & \textbf{Fluency}\\
\cline{2-7}
& \multicolumn{2}{|c|}{Classifier $\uparrow$} & Lexical Scoring $\uparrow$ & \multicolumn{2}{c|}{BLEU $\uparrow$} & \multirow{2}{*}{Perplexity $\downarrow$}\\
& Sentiment & Formality & Formality & \textit{-self} & \textit{-ref} & \\
\hline
Cascaded Style Transformer & \multirow{2}{*}{$72.17$} & \multirow{2}{*}{$64.08$} & \multirow{2}{*}{$81.29$} & \multirow{2}{*}{$0.6066$} & \multirow{2}{*}{$0.3479$} & \multirow{2}{*}{$8.8657$}\\
\cite{dai-etal-2019-style} & &  & & & & \\
Adapted Rewriting LM & \multirow{2}{*}{$52.59$} & \multirow{2}{*}{$36.39$} & \multirow{2}{*}{$72.21$} & \multirow{2}{*}{$\textbf{0.7917}$} & \multirow{2}{*}{$\textbf{0.4259}$} & \multirow{2}{*}{$6.5963$} \\
\cite{adapting-2020-rewritting} & &  & & & & \\
Cascaded Discriminative LM &  $69.30$ & $48.18$ & $83.02$ & $0.6634$ & $0.3579$ & $6.6846$\\
\textbf{Joint Discriminative LM} & $\textbf{79.78}$ & $\textbf{65.33}$ & $\textbf{85.39}$ & $0.7710$ & $0.4136$ & $\textbf{6.4574}$\\
\hline

\end{tabular}
\caption{\label{table-automatic-eval}
Quantitative Comparison of our proposed approach (Joint Discriminative LM) against Cascaded Style Transformer \cite{dai-etal-2019-style}, Cascaded Discriminative LM method and multi-style transfer using Adapted Rewriting LM \cite{adapting-2020-rewritting}. The upward arrow signifies that higher is better and vice versa. Score of near $100$ on formality lexical scoring imply the transferred text is close in formality to the target corpus. }
\end{table*}

\subsection{Style-awareness of Language Models}
\label{section-lm-experiments}
To evaluate style variation across language models fine-tuned on different styles, we compare the generations of the fine-tuned models. For single-dimensional style evaluation, we generate sentences from models fine-tuned on negative corpus and positive corpus and compare the style accuracy of generated sentences. The style accuracy is evaluated by employing a FastText \cite{joulin2016fasttext} classifier trained on the corresponding style dimension. For instance, the classifier for evaluating sentiment accuracy is trained on sentiment corpus tagged with positive and negative class in IMDB and Yelp data. Table~\ref{table-style-fine-tune} shows the accuracy of sentences generated by a model fine-tuned on style $s_i$ as belonging to the class $s_i$. For both sentiment and formality, the fine-tuned language models are able to generate text faithful to the target style dimension. Thus, we conclude that the language models trained on style $s_i$ are able to capture the essence of the corresponding style reasonably well.

These accuracies are an indication of the style awareness in these fine-tuned LMs. We, therefore, employ the perplexities of these fine-tuned language models to gauge the style of the input text to guide our style transfer model. As discussed in discriminative modeling (Section~\ref{section-LM-disc}), the model fine-tuned with corpus from a certain style is expected to have high perplexity on sentence not from that style and low perplexity otherwise. To this end, we experiment with two models independently fine-tuned on positive and negative corpus. We calculate the perplexity of each of these models on the test corpus from the same style and from the opposite style. As seen in Table~\ref{table-fine-tuning-perplexity}, the perplexity for each model is substantially lower on the same corpus as compared to that on the opposite corpus. This implies that a language model fine-tuned on positive corpus shows higher perplexity for negative sentences and lower for positive sentences and vice versa. This corroborates the effectiveness of these fine-tuned language models to serve as discriminators for training the style transfer module.

\subsection{Evaluation metrics}
\label{section-evaluation}
We measure the performance of our model and the baselines based on the style control, content preservation and fluency. 
To measure the \textbf{accuracy of style transfer}, we train two Fasttext\footnote{https://github.com/facebookresearch/fastText} classifiers independently for sentiment and formality using the train corpus, as described in Section~\ref{section-lm-experiments}. These classifiers have accuracy of $93.74\%$ and $88.95\%$ respectively on test corpus of respective datasets. We note that formality as a style is more intricately designed, so we also check lexical scoring by \newcite{brooke-etal-2010-lexical} to evaluate formality, which uses a formality lexicon to assign formality score between $-1$ (informal) and $1$ (formal) to each word and averages it. We scale these scores between $0$--$100$, where higher ($100$) lexical score signifies formal style and lower $(0)$ score signifies informal style. For informal target style, we report lexical score as $100-n$, so that a higher average lexical score signifies a better transfer for either polarity. 

To measure \textbf{content preservation} on transfer, we calculate the BLEU score \cite{papineni-etal-2002-bleu} between the transferred sentence and the input sentence \textit{(self-BLEU)}. Besides this, we also calculate BLEU score between the transferred sentence generated by our model and the corresponding human reference transferred sentence, available for GYAFC and Yelp corpus \textit{(ref-BLEU)}. Since both these corpus account for transfer across only one style dimension each, the provided references are only partial indication of expected outcome. This is also apparent from low ref-BLEU scores for our model as well as baselines. Since, the results are presented on aggregated dataset from both these style dimensions, this evaluation is still able to provide reasonable indication of content preservation.

To measure the \textbf{fluency} of the text, we calculate perplexity assigned to the generated text sequence by a language model trained on the train corpus, as is standard in style transfer literature \cite{dai-etal-2019-style,subramanian-2018-multiple-attribute}. The perplexity is the measure of log likelihood of the generated sentence on the language model. A lower perplexity is indicative of a more fluent sentence. We use a generative transformer-based language model trained on the dataset combined from two styles.

\begin{table*}[th]
\centering
\small
\begin{tabular}{c p{4cm}p{4cm}p{4cm}}
\hline
\multirow{2}{*}{\textbf{Target style}} & \multicolumn{1}{c}{\multirow{2}{*}{\textbf{Source sentence}}} & \multicolumn{2}{c}{\textbf{Transferred Sentence}} \\
& & \multicolumn{1}{c}{\textbf{Style Transformer}} & \multicolumn{1}{c}{\textbf{Our model (multi-style)}} \\
\hline
\\
Positive+Formal & That’s not funny. I don’t think she'll \underline{like it}. & So funny movie. I really like it. & That was very funny. I am sure she will \textcolor{blue}{\textbf{appreciate it}}. \\\\
& Give your brother some money and \underline{tell him to take a hike}. & Just give your brother some time and \textcolor{red}{it will be good again}. & Give your brother some money and \textcolor{blue}{\textbf{request him to leave}}.
\\\\
Negative+Formal & An intelligent, rewarding film that I look forward to watching again. & ludicrous, shallow film that look forward to watching again. & An unintelligent, poor film that \textcolor{blue}{\textbf{I would not look forward}} to watching again.
 \\\\
& \underline{super friendly staff}, quick service and amazing and simple food was done right! & \textcolor{red}{says wait staff}, quick not amazing before overcooked food done were okay. & \textcolor{blue}{dirty staff} and slow service and simple food was not done right. 
\\\\
Positive+Informal & You need to separate the bad thing and move on. & need to the great thing and move on. & You need to enjoy the good stuff and move on.
\\\\
& The evening \underline{started out slow}. & The evening spent in \textcolor{red}{professional show}. & The evening \textcolor{blue}{\textbf{began} amazing}.
\\\\
Negative+Informal & \underline{Great food recommendations} steak and tuna were both great. & \textcolor{red}{terrible food 9am steak} and were both terrible. & \textcolor{blue}{Disappointing food recommendations} steak and tuna were horrible.
 \\\\
& \underline{That person} in hilarious. & \textcolor{red}{You person} in worse! & \textcolor{blue}{\textbf{That guy}} in so boring.
\\
\hline
\end{tabular}
\caption{\label{table-qualitative}
Qualitative results for transfer to different target style combination across different models. (Different colors highlight the transferred segments corresponding to underlined input sentence; Text in bold highlights adherence to target \textit{formality} in text generated by our model.)}
\end{table*}

\begin{table*}[t]
\centering
\small
\begin{tabular}{c|cc|c|c|cc}
\hline
\multirow{2}{*}{Model} & \multicolumn{2}{c|}{Style Accuracy} & Content & \multirow{2}{*}{Fluency} &  Transfer\\
 & Sentiment & Formality & Preservation & & Quality\\
\hline
Cascaded Style Transformer & \multirow{2}{*}{$3.5909$} & \multirow{2}{*}{$2.7424$} & \multirow{2}{*}{$3.2803$} & \multirow{2}{*}{$2.7424$} & \multirow{2}{*}{$2.9318$} \\
\cite{dai-etal-2019-style} & & & & & \\
Joint Discriminative LM & \multirow{2}{*}{$\textbf{3.8561}$} & \multirow{2}{*}{$\textbf{3.0379}$} & \multirow{2}{*}{$\textbf{4.1061}$} & \multirow{2}{*}{$\textbf{4.1894}$} & \multirow{2}{*}{$\textbf{4.1091}$} \\
(Our Model) & & & & & \\
\hline

\end{tabular}
\caption{\label{table-human-eval}
Results for Human Evaluation across different metrics. Each value represents the average of rating between 1 (Very bad) and 5 (Very good).}
\end{table*}

\subsection{Automatic Evaluation}
\newcite{dai-etal-2019-style} use transformer-based model (\textit{Style Transformer}) for single-dimensional style transfer. We train two independent Style Transformer models for sentiment and formality transfer and then perform transfer one after another to compare results with our model. We term this as Cascaded Style Transformer setup. The Style Transformer model is shown to have state-of-the-art performance in single-dimensional style transfer; thus it provides an estimate of the performance of sequential single style transfer.
We also experiment with Adapted Rewriting LM \cite{adapting-2020-rewritting} as another baseline. Their work on style rewriting to match author-specific style does not require explicit annotations for the various aspects that constitutes an author's style, but is based on the assumption that the training corpus reflects the target style. In this context, we train their framework on the mixture of data from the respective target styles and report the performance. These are the closest baselines to our proposed approach, since other works dealing with multi-style transfer assume presence of jointly annotated dataset, which is a stronger assumption that we aim to relax. 
In addition to our proposed model with multiple style transfer, we also train our encoder-decoder architecture with single discriminative LM for one style at a time and perform two stage transfer, similar to one with Cascaded Style Transformer \cite{dai-etal-2019-style} setup.

The results in Table~\ref{table-automatic-eval} show that our model achieves better style control than the Cascaded Style Transformer \cite{dai-etal-2019-style} as well as the joint transfer using \newcite{adapting-2020-rewritting} for both sentiment and formality. As seen in Table~\ref{table-automatic-eval}, cascaded style transfer models perform poorly on content preservation. This is because transferring style one after other leads to huge loss in content, thus both the two-stage models score lower on content preservation metrics, both w.r.t. the input text and the reference transferred text. This demonstrates the advantage of using single model to control for multiple styles. The effect can also be observed in Table~\ref{table-qualitative} which demonstrates qualitative results for Cascaded Style Transformer model and our model. We can see in many cases content loses the underlying meaning of source sentence during the two-stage transfer, whereas our model is able to retain original meaning of the sentence well, corroborating the findings of automatic evaluation.
Among the cascaded models, the Discriminative LM scores marginally better on content preservation than the Style Transformer model. We attribute this to initialization with the same pre-trained LM resulting in shared content space in the underlying single style transfer models. However, due to independent training of the two single style transfer models, they are not able to model interplay between these styles and hence perform worse on style control than our proposed model trained jointly on multiple styles.

Our model also scores better on fluency, as seen in Table~\ref{table-automatic-eval}. This is also apparent from the examples in Table~\ref{table-qualitative}, where sentences generated by Cascaded Style Transformer are much less coherent. Qualitative experiments also highlight the ability of our model to incorporate intricacies of \textit{formality} stylistic dimension (shown in bold) better than the Cascaded Style Transformer model. Among single step transfer models (\newcite{adapting-2020-rewritting} and our proposed approach), we note that content preservation is marginally better for \newcite{adapting-2020-rewritting}'s model, however, our model is able to yield much better style transfer owing to feedback on style control by multiple discriminators. 

\subsection{Human evaluation}
To augment automatic evaluation results, we conduct a human study to evaluate the model outputs across various dimensions such as content preservation, style control, fluency, and overall transfer quality. Based on comparable style control in Cascaded Style Transformer and our proposed approach on automatic metrics, we compare the transfer quality across these two models by a small-scale human study. We select $40$ sentences, with $10$ examples from each combinations of sentiment and formality as target style, and collect annotations from $4$--$5$ participants for each example.
Out of resulting annotations, more than $85\%$ annotations favoured our results over baseline. The average participant rating across different dimensions is shown in Table~\ref{table-human-eval}. We test the statistical signifi
With $\alpha=0.05$, the preferences indicated in human study are significant across all metrics.
These results are in line with our automatic evaluations and add confidence to the efficacy of our proposed approach in achieving style transfer across multiple dimensions.

\section{Conclusion and Future Work}
We propose an approach to extend currently existing style transfer work to multiple style setting without imposing any extra constraints on availability of dataset. Our method makes use of disjoint corpus from separate styles to enable one step transfer across multiple target styles. We exploit multiple discriminative language models with an encoder-decoder framework, all emerging from large transformer-based language models pre-trained on Masked Language Modeling objective and fine-tuned separately for transfer and discriminative purposes. We show that unified single step transfer approach is able to achieve better transfer while offering much better content preservation which is paramount to any style transfer task. 

Further improvements are in scope for adding modularity to the proposed transfer module. In the current setup, each version of model is trained for a specific combination of target style(s). The utility of such a model increases manifold with added ease of transfer across multiple style combinations within a single model. This could be attempted by employing a controlled language model as a unified discriminator for multiple styles, which would be the subject of further research. 

\paragraph{Ethics Statement.} We recognise the ethical implication of employing large language models trained on data infused with unchecked biases. As with any generative task, style transfer too suffers from the potential misuse for fact distortion, plagiarism and more. The paper aims at establishing academic utility of proposed framework. To meet ethical standards, this solution has to coupled with strict misrepresentation, offensiveness and bias checks.

\bibliography{anthology,custom}
\bibliographystyle{acl_natbib}

\end{document}